\pdfoutput=1
\documentclass[11pt]{article}
\usepackage[]{acl}

\usepackage{times}
\usepackage{latexsym}
\usepackage{adjustbox}
\usepackage{multirow}

\usepackage{amsmath}
\usepackage{graphicx}
\usepackage{float}
\usepackage{hyperref}
\usepackage{booktabs}

\usepackage[T1]{fontenc}

\usepackage[utf8]{inputenc}

\usepackage{microtype}

\title{Conditional Bilingual Mutual Information Based Adaptive Training for Neural Machine Translation}

\author{Songming           Zhang\textsuperscript{1}\thanks{ \ \ Equal contribution. Work was done when Songming were interning at Pattern Recognition Center, WeChat AI, Tencent Inc, China.},
Yijin Liu\textsuperscript{2}\footnotemark[1] , 
Fandong Meng\textsuperscript{2},
\textbf{Yufeng Chen}\textsuperscript{1}\thanks{ \ \ Yufeng Chen is the corresponding author.}, \\ 
\textbf{Jinan Xu}\textsuperscript{1},
\textbf{Jian Liu}\textsuperscript{1} and \textbf{Jie Zhou}\textsuperscript{2}\\
\textsuperscript{1}Beijing Key Lab of Traffic Data Analysis and Mining, \\
Beijing Jiaotong University, Beijing, China \\
\textsuperscript{2}Pattern Recognition Center, WeChat AI, Tencent Inc, China \\
\texttt{\{zhangsongming,chenyf,jaxu,jianliu\}@bjtu.edu.cn}, \\
\texttt{\{yijinliu,fandongmeng,withtomzhou\}@tencent.com} \\}

\begin{document}
\maketitle
\begin{abstract}
% corresponding to
Token-level adaptive training approaches can alleviate the token imbalance problem and thus improve neural machine translation, 
through re-weighting the losses of different target tokens based on specific statistical metrics ({\em e.g.,} token frequency or mutual information).
% certain  autoregressively
Given that standard translation models make predictions on the condition of previous target contexts, we argue that the above statistical metrics ignore target context information and may assign inappropriate weights to target tokens. 
% excessive unacceptable
% However, it is non-trivial to take target contexts into these statistical metrics, 
While one possible solution is to directly take target contexts into these statistical metrics, the target-context-aware statistical computing is extremely expensive, and the corresponding storage overhead is unrealistic.
To solve the above issues, we propose a target-context-aware metric, named conditional bilingual mutual information (CBMI), which makes it feasible to supplement target context information for statistical metrics.
% , {\em i.e.,} mutual information in this paper. 
% which measures the dependencies between target tokens and source sentences on the condition of previous target contexts. 
Particularly, our CBMI can be formalized as the log quotient of the translation model probability and language model probability by decomposing the conditional joint distribution.
% Furthermore, our CBMI can be simply formalized as the quotient of the translation probability and the language model probability through the decomposition of conditional joint distribution. 
Thus CBMI can be efficiently calculated during model training without any pre-specific statistical calculations and large storage overhead.
% Furthermore, we propose an effective and efficient adaptive training approach based on the token- and sentence-level CBMI measures.
Furthermore, we propose an effective adaptive training approach based on both the token- and sentence-level CBMI.
Experimental results on WMT14 English-German and WMT19 Chinese-English tasks show our approach can significantly outperform the Transformer baseline and other related methods.

\end{abstract}
\vspace{-7pt}

\begin{figure}[t]
    \centering
    \includegraphics[width=\linewidth]{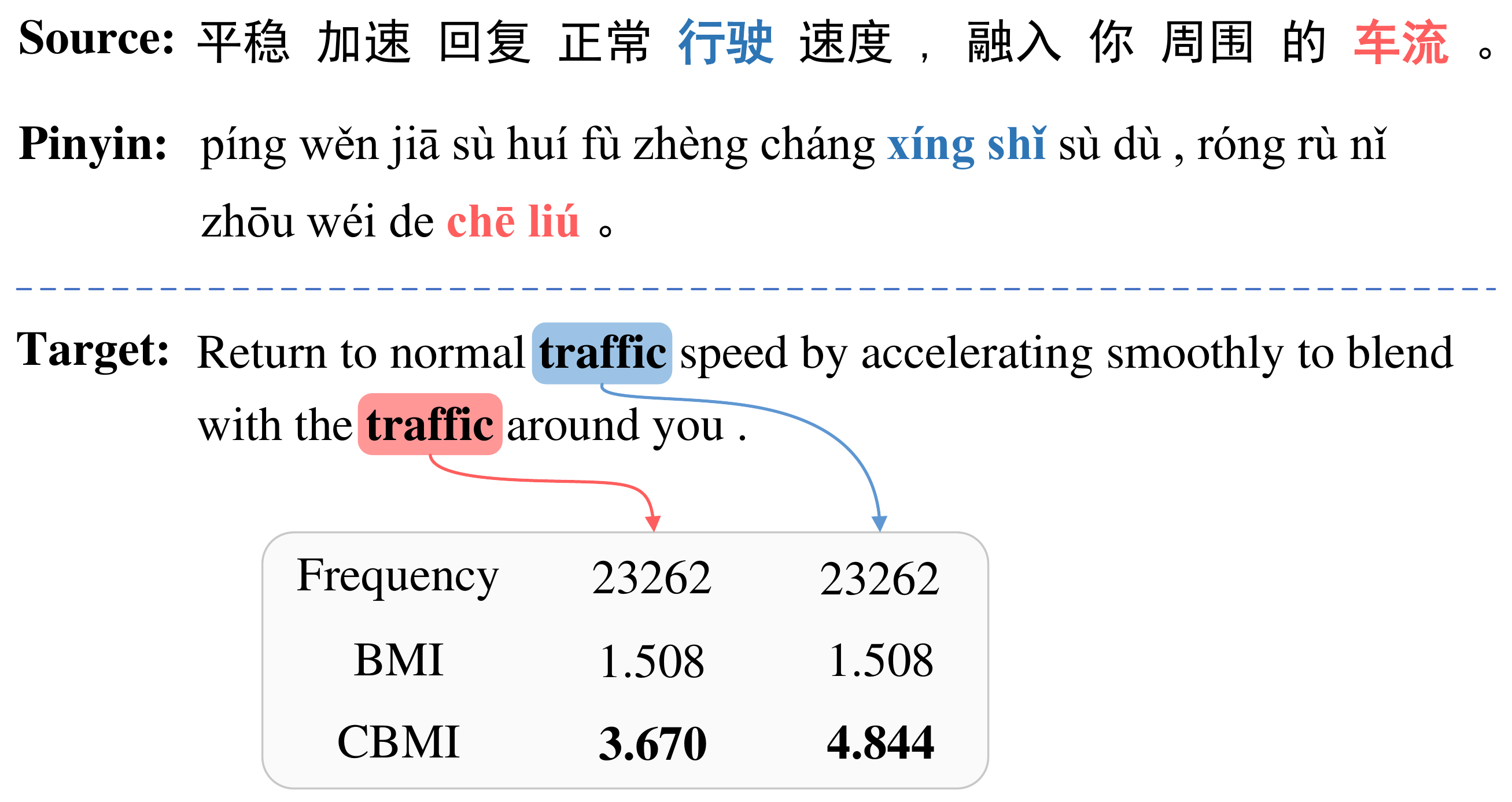}
    \vspace{-10pt}
    \caption{An example from the WMT19 Zh-En training set. Despite the different mappings from the source sentence, existing target-context-free metrics ({\em i.e.,} frequency and BMI) equally assess the two `\textit{traffic}' tokens, while our CBMI can distinguish the different dependencies of the two tokens on the source sentence with the guidance of target contexts.}
    % and assign different training weights to the two tokens.}
    \label{fig:case}
    \vspace{-7pt}
\end{figure}

\section{Introduction}

Neural machine translation (NMT) \cite{bahdanau2014neural,convs2s,transformer,meng2019dtmt,conf_ss,dec_ss} has made remarkable achievements in recent years. 
Generally, NMT models are trained to maximize the likelihood of the next target token given ground-truth tokens as inputs \cite{johansen1990maximum, teacher_forcing_2016}. 
% Generally, NMT models are optimized through maximum likelihood estimation (MLE) \cite{johansen1990maximum, teacher_forcing_2016}. 
Due to the token imbalance phenomenon in natural language \cite{zipf_human_1949}, for an NMT model, the learning difficulties of different target tokens may be various. However, the vanilla NMT model equally weights the training losses of different target tokens, irrespective of their difficulties.
% It is preferable to distinguish the contribution of different tokens according their .

Recently, various adaptive training approaches \cite{gu-etal-2020-token,xu-etal-2021-bilingual} have been proposed to alleviate the above problem for NMT.
% balance the training difficulty of the NMT models. 
Generally, these approaches re-weight the losses of different target tokens based on specific statistical metrics. For example, \citet{gu-etal-2020-token} take the token frequency as an indicator and encourage the NMT model to focus more on low-frequency tokens.
% re-weight the training loss of different target tokens according to their frequencies.  
\citet{xu-etal-2021-bilingual} further propose the bilingual mutual information (BMI) to measure the word mapping diversity between bilinguals, and down-weight the tokens with relatively lower BMI values.

Despite their achievements, there are still limitations in these adaptive training approaches.
Given that the standard translation model autoregressively makes predictions on the condition of previous target contexts, we argue that the statistical metrics used in the above approaches ignore target context information and may assign inaccurate weights for target tokens.
Specifically, although existing statistical metrics can reflect complex characteristics of target tokens ({\em e.g.,} mapping diversity), they fail to model how these properties vary across different target contexts. 
Secondly, for the identical target tokens in different positions of a target sentence ({\em e.g.,} two `\textit{traffic}' tokens in the Figure \ref{fig:case}), they may be mapped from different source-side tokens, but such target-context-free metrics cannot distinguish the above different mappings. 
% with the gradual increase of target contextual information during translation,
% modeling such target context information can help the NMT model better learn polysemy (e.g., the same word in different positions of a target sentence).
In summary, it is necessary to incorporate target context information into the above statistical metrics.
One possible solution is to directly take target context information into account and conduct target-context-aware statistical calculations.
But in this way, the calculation cost and storage overhead will become huge and unrealistic\footnote{Take the vanilla BMI \cite{xu-etal-2021-bilingual} as an example, to process the raw WMT14 En-De training data (about 1.5GB), it takes about 12 CPU hours and 2GB disk storage to save the BMI values. To make matters worse, the cost will increase dozens of times in target-context-aware statistical calculations.}.
Therefore, it is non-trivial to design a suitable target-context-aware statistical metric for adaptive training in the field of NMT.

In this paper, we aim to address the above issues in adaptive training methods.
Firstly, we propose a novel target-context-aware metric, named \textbf{C}onditional \textbf{B}ilingual \textbf{M}utual \textbf{I}nformation (CBMI), to measure the importance of different target tokens by their dependence on the source sentence. 
% measure the learning difficulties of different target tokens. 
Specifically, we calculate CBMI by the mutual information between a target token and its source sentence on the condition of its target contexts. 
With the aid of target-context-aware calculations, CBMI can easily model the various characteristics of target tokens under different target contexts, and of course can distinguish identical target tokens with different source mappings.
Regarding the computational efficiency, through decomposing the conditional joint distribution in the aforementioned mutual information, our CBMI can be formalized as the log quotient of the translation model probability and language model probability\footnote{The detailed derivation process is shown in Equation \eqref{eq:derive}. Please note that the language model is only used during training and thus does not affect the inference speed. }. Therefore, CBMI can be efficiently calculated during model training without any pre-specific statistical calculations and huge storage overhead, which makes it feasible to supplement target context information for statistical metrics.
% Besides efficient calculations, our CBMI provides a more suitable assessment of the token importance when compared with existing target-context-free metrics. 
% For example, as shown in the Figure \ref{fig:case}, existing target-context-free metrics ({\em i.e.,} frequency and BMI) equally treat the two `\textit{traffic}' tokens, despite their different target contexts.
% In contrast, our CBMI can distinguish the two tokens according to their target contexts and thus provide more suitable assessments.
Subsequently, we design an adaptive training approach based on both the token- and sentence-level CBMI, which dynamically re-weights the training losses of the corresponding target tokens. 

We evaluate our approach on the WMT14 English-German and WMT19 Chinese-English translation tasks. Experimental results on both datasets demonstrate that our approach can significantly outperform the Transformer baseline and other adaptive training methods.
% Further analyses reveal that the measurement of CBMI also highly relates to the translation adequacy and our approach can improve translation adequacy meanwhile maintaining fluency.
Further analyses reveal that CBMI can also reflect the adequacy of translation, and our CBMI-based adaptive training can improve translation adequacy meanwhile maintain fluency.
The main contributions of this paper can be summarized as follows:
\begin{itemize}
    \vspace{-2pt}
    \item We propose a novel target-context-aware metric, named CBMI, which can reflect the importance of target tokens for NMT models.
    Theoretical analysis and experimental results show that CBMI is computationally efficient, which makes it feasible to complement target context information in statistical metrics.
    \vspace{-2pt}
    % than existing statistical metrics 
    \item We further propose an adaptive training approach based on both the token- and sentence-level CMBI, which dynamically re-weights the training losses of target tokens.
    \vspace{-2pt}
    \item Further analyses show that CBMI can also reflect the adequacy of translation, and CBMI-based adaptive training can improve translation adequacy meanwhile maintain fluency\footnote{The code is publicly available at: \url{https://github.com/songmzhang/CBMI}.}.
    % \vspace{-3pt}
    
\end{itemize}

% \vspace{-3pt}
\section{Background}
% \vspace{-3pt}
\subsection{Neural Machine Translation}
An NMT model is designed to translate a source sentence with $M$ tokens $\mathbf{x}=\{x_1, x_2, \ldots, x_M\}$ into a target sentence with $N$ tokens $\mathbf{y}=\{y_1, y_2, \ldots, y_N\}$ by predicting the probability of each target token:
\begin{equation}
    P(\mathbf{y}|\mathbf{x};\theta)= \prod_{j=1}^N p(y_j|\mathbf{y}_{<j},\mathbf{x};\theta)
\end{equation}
where $j$ is the index of each time step, $\mathbf{y}_{<j}$ is the target-side previous context for $y_j$, and $\theta$ is the model parameter.

During training, NMT models are generally optimized with the cross-entropy (CE) loss:
\begin{equation} \label{eq:ce_loss}
    \mathcal{L}_{\rm CE}(\theta)= -\sum_{j=1}^N\log p(y_j|\mathbf{y}_{<j},\mathbf{x};\theta)
\end{equation}

During inference, NMT models predict the probabilities of target tokens in an auto-regressive mode and generate hypotheses using heuristic search algorithms like beam search \cite{beam_search}.
% \vspace{-5pt}

\subsection{Token-level Adaptive Training for NMT}
Token-level adaptive training aims to alleviate the token imbalance problem for NMT models by re-weighting the  training losses of target tokens. 
How to design a suitable weight adjustment strategy matters, which is we aim to improve in this paper. 
% Generally, existing approaches take statistical metrics as measurement, {\em e.g.,}, token frequency and mutual information.
Formally, for the $j$-th target token and its adaptive weight $w_{j}$, the standard cross-entropy loss in Equation \eqref{eq:ce_loss} is expanded to the following formula:
\vspace{-7pt}
\begin{align}
    % w_j&=\Phi({\rm measurement}_j) \\
    \mathcal{L}_{\rm ada}(\theta)&=-\sum_{j=1}^N w_j\log p(y_j|\mathbf{y}_{<j},\mathbf{x};\theta) \label{eq:general_weight_adjust}
    \vspace{-3pt}
\end{align}
% where $w_j$ is the adaptive training weight for $y_j$.
% \vspace{-10pt}

\subsection{Mutual Information for NMT}

Mutual information (MI) is a general metric in information theory \cite{information_theory}, which measures the mutual dependence between two random variables $a$ and $b$ as follows\footnote{We use the point-wise MI here instead of the original expectation form, since we aim to calculate the mutual information between individual samples in this paper.}:
\vspace{-5pt}
\begin{equation} \label{eq:mi}    
    {\rm MI}(a;b)=\log\left(\frac{p(a,b)}{p(a)\cdot p(b)}\right)
\vspace{-5pt}
\end{equation}

\citet{xu-etal-2021-bilingual} propose token-level bilingual mutual information (BMI) to measure the word mapping diversity between bilinguals and further conduct BMI-based adaptive training for NMT. The BMI is formulated as:
\vspace{-5pt}
\begin{equation} \label{eq:bmi}
    {\rm BMI}(\mathbf{x};y_j)= \sum_{i=1}^{|\mathbf{x}|} \log \left(\frac{f(x_i,y_j)}{f(x_i)\cdot f(y_j)}\right)
    \vspace{-5pt}
\end{equation}
% offline
where $f(\cdot)$ is an word frequency counter.
Although BMI can reflect the bilingual mapping properties to some extent, it cannot correspondingly vary with the target context. However, simply introducing target-context-aware calculations into BMI would make the above statistical calculations unrealistic.

% , and needs additional storage overhead to save the BMI values.
% Different from BMI, our CBMI does not rely on the time-consuming pre-processing procedure since it can be calculated by models in real time during the training stage.
% \vspace{-5pt}
\section{Approaches}
\begin{figure}[t]
    \centering
    \includegraphics[width=0.9\linewidth]{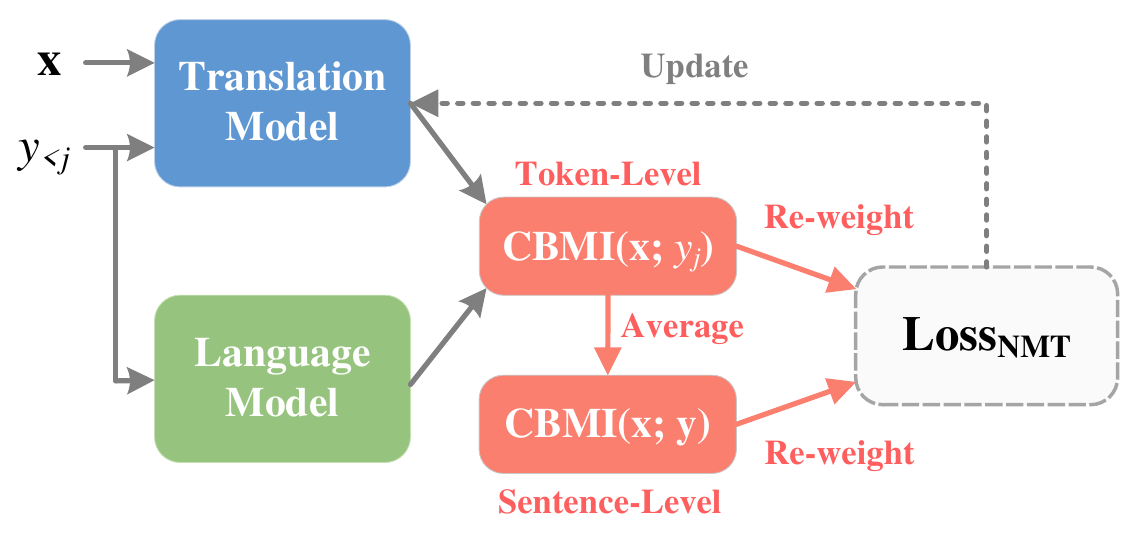}
    \vspace{-2pt}
    \caption{Overview of the training process of our method. For the target token $y_j$, we calculate its token-level CBMI by the translation model and the language model, and average all the token-level CBMI values in a sentence into the sentence-level CBMI. Then the two CBMI values with different granularities are combined to form the final training loss weight of the token $y_j$.}
    \label{fig:overview}
    \vspace{-2pt}
\end{figure}

% \vspace{-5pt}
In this section, we first introduce the definition of CBMI (Section \ref{sec:cbmi}). Then, we illustrate how to adjust the weights for the training losses of target tokens based on the token- and the sentence-level CBMI (Section \ref{sec:weight}).
% Then we elaborate how to adjust training weights (Section \ref{sec:weight}) and how to select prior distributions to guide training (Section \ref{sec:selection}) in our CBMI-based objective. 
% Finally, we formulate the training losses of these strategies and the final objective (Section \ref{sec:train}). 
Figure \ref{fig:overview} shows the overall training process of our approach. 
% \vspace{-5pt}
\subsection{Definition of CBMI} \label{sec:cbmi}
% \vspace{-5pt}

As mentioned above, it is necessary to incorporate target context information into the statistical metrics ({\em e.g.,} BMI) for adaptive training.
However, it is impractical to directly conduct target-context-aware statistical computations due to the expensive computational costs and storage overhead.
In this paper, we propose a new target-context-aware metric, named conditional bilingual mutual information (CBMI), to solve the above issues.
% Unfortunately, directly appending the target context to these metrics is unfeasible due to the massive statistical computing. To tackle these problems, we propose a target-context-aware metric, named conditional bilingual mutual information (CBMI). 
Specifically, CBMI is calculated by the mutual information between each target token and its source sentence under the condition of previous target context.
% so that CBMI can measure their mutual dependency on the condition of the target context. 
Formally, the CBMI of a target token $y_{j}$ and its source sentence $\mathbf{x}$ is calculated as follow:
\begin{equation}
    \label{eq:cbmi}
    \resizebox{0.89\linewidth}{!}{
    $
    \begin{split}
        {\rm CBMI}(\mathbf{x};y_j) &= {\rm MI}\left(\mathbf{x};y_j|\mathbf{y}_{<j}\right) \\
        &=\log \left(\frac{p(y_j,\mathbf{x}|\mathbf{y}_{<j})}{p(y_j|\mathbf{y}_{<j})\cdot p(\mathbf{x}|\mathbf{y}_{<j})}\right) \\
    \end{split}
    $
}
\end{equation}

The original CBMI definition presented in the above equation still struggles in computation, thus we further simplify it by decomposing the conditional joint distribution:
% \vspace{-1pt}
\begin{equation}
    \label{eq:derive}
    \resizebox{0.89\linewidth}{!}{
    $
    \begin{split} 
    % \resizebox{\linewidth}{!}{
    {\rm CBMI}(\mathbf{x};y_j) 
    % &= {\rm MI}\left(\mathbf{x};y_j|\mathbf{y}_{<j}\right) \\
    &=\log\left(\frac{p(y_j,\mathbf{x}|\mathbf{y}_{<j})}{p(y_j|\mathbf{y}_{<j})\cdot p(\mathbf{x}|\mathbf{y}_{<j})}\right) \\
    &=\log\left(\frac{p(y_j|\mathbf{x},\mathbf{y}_{<j})\cdot p(\mathbf{x}|\mathbf{y}_{<j})}{p(y_j|\mathbf{y}_{<j})\cdot p(\mathbf{x}|\mathbf{y}_{<j})}\right) \\
    &=\log\left(\frac{p(y_j|\mathbf{x},\mathbf{y}_{<j})}{p(y_j|\mathbf{y}_{<j})}\right) \\
    &=\log\left(\frac{p_{\rm NMT}(y_j)}{p_{\rm LM}(y_j)}\right)
    % }
    \end{split}
    $
    }
    % \vspace{-1pt}
\end{equation}
where $p_{\rm NMT}(y_{j})$ is the probability output by the NMT model, and $p_{\rm LM}(y_{j})$ is the probability output by an additional target-side language model (LM). 
% elaborately/succinctly
In this way, we formalize the complex target-context-aware calculation in Equation \eqref{eq:cbmi} as the log quotient of the NMT probability and LM probability.
Based on the simplified Equation \eqref{eq:derive}, CBMI can be computed in real time during the model training, thus enabling both target-context-aware and efficient computations. 
% Note that the additional target-side LM is only used during training and does not affect the inference speed. 
Considering the massive computation required by existing methods to perform the target-context-aware calculation, the LM in our CBMI only brings a modest computational cost in training and finally leads to better performance.
We will give a detailed comparison of the calculation cost and storage overhead between our CBMI and existing approaches in Section \ref{sec:analysis_cost}.

% According to the above formula, our CBMI can be obtained through efficient calculation by an NMT model and a language model. 
% Compared with the target-context-aware statistical metrics, which will introduce excessive statistical computation to incorporate target-contexts, CBMI achieves the same goal in a more efficient way during training. Therefore, CBMI is a more realistic solution to supplement the target-contextual information and can further provide a more suitable assessment for each target token.
% \vspace{-1pt}
\subsection{CBMI-based Weight Adjustment} \label{sec:weight}
According to the definition, CBMI measures the mutual dependence between a target token and its corresponding source sentence on the condition of its context. 
Namely, target tokens with larger CBMI value rely more on the source-side information and less on the target historical translations, which is exactly in line with the goal of the adequacy translation model. 
Given that current NMT models tend to generate fluent but inadequate translations \cite{enhance_faithfulness,mengqi_margin}, we speculate that making the NMT models pay more attention to target tokens with larger CBMI values can improve translation adequacy and thus improve translation performance.
Furthermore, we observe a phenomenon that if target sentences contain many words with small CBMI values, they generally do not match well with the corresponding source sentences. To alleviate the negative effect of these poorly matched sentence pairs, we average all the token-level CBMI values in a target sentence into a sentence-level CBMI and incorporate it into our approach. 
Consequently, we propose to dynamically adjust the training weight of each target token based on both the token- and sentence-level CBMI. 
For clarity, we use $t$ to mark the `token-level' intermediate variables and $s$ to mark the `sentence-level' ones in the following formulas.

% We connect this characteristic to the translation adequacy of NMT models, which expects higher source information coverage during translation. From this aspect, we speculate that those tokens with higher CBMI contribute more to enhancing the adequacy of NMT models. 
% Since current NMT models tend to generate fluent yet inadequate translations \cite{enhance_faithfulness,mengqi_margin}, we speculate that tokens with higher CBMI are more important for NMT models. For this reason, we regard CBMI as an importance metric for the assessment of target tokens and up-weight the tokens with higher importance measurement.

% To stabilize the distribution of CBMI values in different sentences, we firstly conduct intra-sentence normalization for token-level CBMI. weights for target tokens based on current token-level CBMI. To further compare the importance of target tokens in different sentences, we then calculate sentence-level CBMI to incorporate the sentence information into each token. Finally, the adaptive training weight for the loss of each target tokens are composed of both the token-level and the sentence-level training weight.
% Under the measurement of CBMI, we further explore both token-level and sentence-level CBMI to adaptively adjust the training weight of each token. For clear distinction, we use $t$ to mark the `token-level' intermediate variables and $s$ to mark the `sentence-level' ones in the following formulas.
% \vspace{-5pt}
\paragraph{Token-Level CBMI.} 
The token-level CBMI can reflect the importance of target tokens for improving translation adequacy ({\em i.e.}, dependency of the source side information). Thus we amplify the weights of target tokens with larger token-level CBMI to make the NMT model pay more attention to them.
% these target tokens.
% From the token-level perspective, target tokens with higher CBMI values in a sentence are considered more important for NMT models. Therefore, we assign larger training weights to amplify the losses of these tokens. 
Particularly, to reduce the variances and stabilize the distribution of the token-level CBMI in each target sentence, we firstly conduct intra-sentence normalization for the token-level CBMI ${\rm CBMI}^t(\mathbf{x};y_j)$:
% \vspace{-0.5cm}
\begin{equation}
    \resizebox{.85\hsize}{!}{${\rm CBMI}_{norm}^t(\mathbf{x};y_j)=({\rm CBMI}^t(\mathbf{x};y_j) - \mu^{t}) / \sigma^{t}$} \label{eq:token_norm}
\end{equation}
where $\mu^{t}$, $\sigma^{t}$ represent the mean values and the standard deviations of ${\rm CBMI}^t(\mathbf{x};y_j)$ in each target sentence.

Then we scale the normalized CBMI value ${\rm CBMI}^t_{norm}(\mathbf{x};y_j)$ to obtain the token-level training weight for $y_j$:
% \footnote{We limit the minimum weights to 0 and the mean weights to 1}
% \vspace{-0.5cm}
\begin{equation}
    w_j^{t}={\rm max}\{0, scale^{t} \cdot {\rm CBMI}^t_{norm}(\mathbf{x};y_j) + 1\} \label{eq:wt}
\end{equation}
where $scale^t$ is a hyperparameter that controls the effect of ${\rm CBMI}^t_{norm}(\mathbf{x};y_j)$.

\paragraph{Sentence-level CBMI.} 
% From the sentence-level perspective, the bilingual mutual dependencies in different sentence pairs are also various, which is embodied in their various matching degrees. 
We average all the token-level CBMI values in a target sentence to form the sentence-level CBMI, which can further reflect the matching degree between the bilingual sentences in a sentence pair.
To alleviate the negative effect of poorly matched sentence pairs and encourage the NMT model focus on well-matched sentences pairs, we up-weight the sentence pairs with larger sentence-level CBMI values and down-weight those sentence pairs with smaller sentence-level CBMI values.
% We observe that if target sentences in the training data contain many words with small CBMI value, they generally do not match well with the corresponding source sentences. To alleviate the negative effect of these poorly matched sentence pairs, we average all the token-level CBMIs in a target sentence into a sentence-level CBMI, which is also taken into account when designing our adaptive weighting approach. 
% Thus we introduce sentence-level CBMI to measure the dependencies and accordingly amply the contribution of those well-matched sentences during training.
Specifically, the sentence-level CBMI between the source sentence $\mathbf{x}$ and the target sentence $\mathbf{y}$ can be derived from Equation \eqref{eq:mi} and represented as the arithmetic average of token-level CBMI values\footnote{We divide the original sentence CBMI with its corresponding sentence length to reduce its variance.}:
\vspace{-2pt}
\begin{equation}
    \label{eq:sentence_cbmi}
    \resizebox{0.85\linewidth}{!}{
    $
    \begin{split}
        {\rm CBMI}^s(\mathbf{x};\mathbf{y})&=\frac{1}{|\mathbf{y}|}\log\left(\frac{p(\mathbf{x},\mathbf{y})}{p(\mathbf{x})\cdot p(\mathbf{y})}\right) \\
    &=\frac{1}{|\mathbf{y}|}\log\left(\frac{p(\mathbf{y}|\mathbf{x})}{p(\mathbf{y})}\right) \\
    &=\frac{1}{|\mathbf{y}|}\log\left(\frac{\prod_{j}p(y_j|\mathbf{x},\mathbf{y}_{<j})}{\prod_{j}p(y_j|\mathbf{y}_{<j})}\right) \\
    &=\frac{1}{|\mathbf{y}|}\sum_{j}\log\left(\frac{p(y_j|\mathbf{x},\mathbf{y}_{<j})}{p(y_j|\mathbf{y}_{<j})}\right) \\
    &=\frac{1}{|\mathbf{y}|}\sum_{j} {\rm CBMI}^t(\mathbf{x};y_j)
    \end{split}
    $
    }
    \vspace{-2pt}
\end{equation}

Similarly, we conduct inter-sentence normalization for ${\rm CBMI}^s(\mathbf{x};\mathbf{y})$:
\vspace{-2pt}
\begin{equation}
    \resizebox{.85\hsize}{!}{${\rm CBMI}^s_{norm}(\mathbf{x};\mathbf{y})=({\rm CBMI}^s(\mathbf{x};\mathbf{y}) - \mu^{s}) / \sigma^{s}$}
    \label{eq:sentence_norm}
    % \vspace{-3pt}
\end{equation}
where $\mu^{s}$, $\sigma^{s}$ represent the mean values and the standard deviations of ${\rm CBMI}^s(\mathbf{x};\mathbf{y})$ in each mini-batch during training.

Subsequently, we also scale ${\rm CBMI}^s_{norm}(\mathbf{x};\mathbf{y})$ in Equation \eqref{eq:sentence_norm} with another hyperparameter $scale^{s}$ to obtain the sentence-level training weight:
% \vspace{-2pt}
\begin{equation}
    \resizebox{.85\hsize}{!}{$w^{s}={\rm max}\{0, scale^{s} \cdot {\rm CBMI}^s_{norm}(\mathbf{x};\mathbf{y}) + 1\}$}
     \label{eq:ws}
\end{equation}
% \vspace{-5pt}
\paragraph{Final Loss Weight.} In our adaptive training approach, for the target token $y_j$, its final loss weight $w_j$ in Equation \eqref{eq:general_weight_adjust} is the multiplication of the above two weights in Equation \eqref{eq:wt} and \eqref{eq:ws}:
% \vspace{-5pt}
\begin{equation} \label{eq:weight}
    w_j = w_j^{t} \cdot w^{s}
\end{equation}

\vspace{-3pt}
\section{Experiments}

\subsection{Datasets}
We conduct experiments on two large-scale WMT tasks, {\em i.e.,} the WMT14 English to German (En-De) and WMT19 Chinese to English (Zh-En). For the En-De task, the training set contains 4.5M sentence pairs. The validation set and test set are newstest2013 and newstest2014, respectively. For the Zh-En task, the training set totally contains 20M sentence pairs and the validation set and test set are newstest2018 and newstest2019, respectively. 
Following previous work, we share the vocabulary for the En-De task and segment words into subwords using byte pair encoding (BPE) \cite{sennrich-etal-2016-neural} with 32k merge operations for both datasets.
% \vspace{-5pt}
\subsection{Implementation Details}
\paragraph{Training.}We implement baselines and our approach under Transformer$_{base}$ and Transformer$_{big}$ settings based on the open-source toolkit fairseq \cite{ott2019fairseq} with mixed precision \cite{ott-etal-2018-fp16}. We train all the translation models with the cross-entropy loss for 100k steps, and further finetune them with different adaptive training objectives for another 200k steps on both tasks.
The target-side language model is a Transformer decoder without the cross-attention modules, which is trained synchronously with the translation model. 
The training data for the language model is the target-side monolingual data from the NMT training set.
% The target-side language model keeps the same architecture with the Transformer decoder except for the cross-attention modules, and has no parameter sharing with the NMT model in our approach. 
All the experiments are conducted on 8 NVIDIA Tesla V100 GPUs, and each batch on each GPU contains approximately 4096 tokens. We use Adam optimizer \cite{article} with 4000 warmup steps to optimize models. 
More training details are listed in Appendix \ref{sec:appendix_config}.

In our experiments, we have not been able to bring further improvement to our approach through simply enhancing the language model. Our conjecture is that stronger language models will generate sharper distribution, and will increase the variances of CBMI values when used as the denominator, resulting in detriment for NMT model training. We will leave this for the future work.

\paragraph{Evaluation.}During inference, we set beam size to 4 and length penalty to 0.6 for both tasks. We use \textit{multibleu.perl} to calculate case-sensitive BLEU for WMT14 En-De and \textit{SacreBLEU}\footnote{SacreBLEU hash:\qquad\quad BLEU+case.mixed+numrefs.1\\+smooth.exp+tok.13a+version.1.5.1.} to calculate case-sensitive BLEU for WMT19 Zh-En. We use the paired bootstrap resampling methods \cite{koehn-2004-statistical} for the statistical significance test.
% \vspace{-5pt}
\subsection{Hyperparameter Experiments.}
% \vspace{-5pt}
In this section, we introduce the hyperparameter settings of our approach according to the performance on the validation set of the WMT14 En-De dataset, and we share the same hyperparameter settings with the WMT19 Zh-En dataset.
% \vspace{-5pt}

\begin{table*}[t] 
\centering
\resizebox{0.85\linewidth}{!}{
\begin{tabular}{l|c|c}
\bottomrule
\textbf{Model} \qquad\qquad\qquad\qquad\qquad\qquad & \textbf{WMT14 En$\rightarrow$De} & \textbf{WMT19 Zh$\rightarrow$En}  \\
\hline
\hline
% \textit{Base Setting} & & \\
Transformer$_{base}$ \cite{transformer} {$\dagger$} & 27.30 & --\\
Transformer$_{base}$ \cite{transformer} & 28.10 & 25.36 \\
\quad + Freq-Exponential \cite{gu-etal-2020-token} & 28.43 (+0.33) & 24.99 (-0.37)\\
\quad + Freq-Chi-Square \cite{gu-etal-2020-token} & 28.47 (+0.37) & 25.43 (+0.07)\\
\quad + BMI-adaptive \cite{xu-etal-2021-bilingual} & 28.56 (+0.45) & 25.77 (+0.41)\\
% \quad + BMI-Adaptive \cite{xu-etal-2021-bilingual} {$\dagger$} & 28.53 & 25.19 \\
\quad + Focal Loss \cite{focal_loss} & 28.43 (+0.33) & 25.37 (+0.01)\\
\quad + Anti-Focal Loss \cite{raunak-etal-2020-long} & 28.65 (+0.55) & 25.50 (+0.14)\\
\quad + Self-Paced Learning \cite{wan-etal-2020-self} & 28.69 (+0.59)  & 25.75 (+0.39) \\
% \quad + D2GPo \cite{Li2020Data-dependent} & 28.24 (+0.22) & $-$ \\
% \quad + Mixed Cross Entropy \cite{pmlr-v139-li21n} & 28.32 (+0.30) & = \\
\quad + Simple Fusion \cite{stahlberg-etal-2018-simple} & 27.82 (-0.28) & 23.91 (-1.45) \\
\quad + LM Prior \cite{baziotis-etal-2020-language} & 28.27 (+0.17) & 25.71 (+0.35) \\
\quad + CBMI-adaptive (ours) & \textbf{\ \ \ 29.01 (+0.91)$^{\ast\ast}$} & \textbf{\ \ \ 26.21 (+0.85)$^{\ast\ast}$}\\
% \quad + CBMI-prior selection (ours) & 28.48 (+0.46)$^\ast$ & $-$\\
% \quad + CBMI-both weight and prior selection (ours) & \textbf{28.94 (+0.92)$^\ast$} & $-$\\
\midrule
% \textit{Big Setting} & & \\
Transformer$_{big}$ \cite{transformer} {$\dagger$} & 28.40  & --\\
Transformer$_{big}$ \cite{transformer} & 29.31  & 25.48 \\
\quad + Freq-Exponential \cite{gu-etal-2020-token} & 29.66 (+0.35) & 25.57 (+0.09) \\
\quad + Freq-Chi-Square \cite{gu-etal-2020-token} & 29.64 (+0.33) & 25.64 (+0.14)\\
\quad + BMI-adaptive \cite{xu-etal-2021-bilingual} & 29.69 (+0.38) & 25.81 (+0.33) \\
\quad + Focal Loss \cite{focal_loss} & 29.65 (+0.34) & 25.54 (+0.06)\\
\quad + Anti-Focal Loss \cite{raunak-etal-2020-long} & 29.72 (+0.41) & 25.64 (+0.16)\\
\quad + Self-Paced Learning \cite{wan-etal-2020-self} & 29.85 (+0.54) & 25.88 (+0.40)\\
% \quad + Simple Fusion \cite{stahlberg-etal-2018-simple} & & \\
% \quad + D2GPo \cite{Li2020Data-dependent} & $-$ & $-$\\
% \quad + Mixed Cross Entropy \cite{pmlr-v139-li21n} & 29.12 (-0.04) & $-$\\
\quad + CBMI-adaptive (ours) & \textbf{\ \,30.12 (+0.81)$^{\ast}$} & \textbf{\ \,26.30 (+0.82)$^{\ast}$}\\
% \quad + CBMI-prior selection (ours) & = & $-$ \\
% \quad + CBMI-both weight and prior selection (ours) & = & - \\
\toprule
\end{tabular}
}
\caption{BLEU scores (\%) on two translation tasks. Each experiment runs over 3 times and we list the mean values and improvements in this table (full results including standard deviations are shown in Appendix \ref{sec:full}). `{$\dagger$}' represents the results taken from the corresponding papers. Results with mark $\ast$/$\ast\ast$ are statistically \cite{koehn-2004-statistical} better than the most related method `BMI-Adaptive' with $p<0.05$ and $p<0.01$.
}

\label{tab:main_results}
\end{table*} 

\begin{figure}[t]
    \centering
    \includegraphics[width=0.88\linewidth]{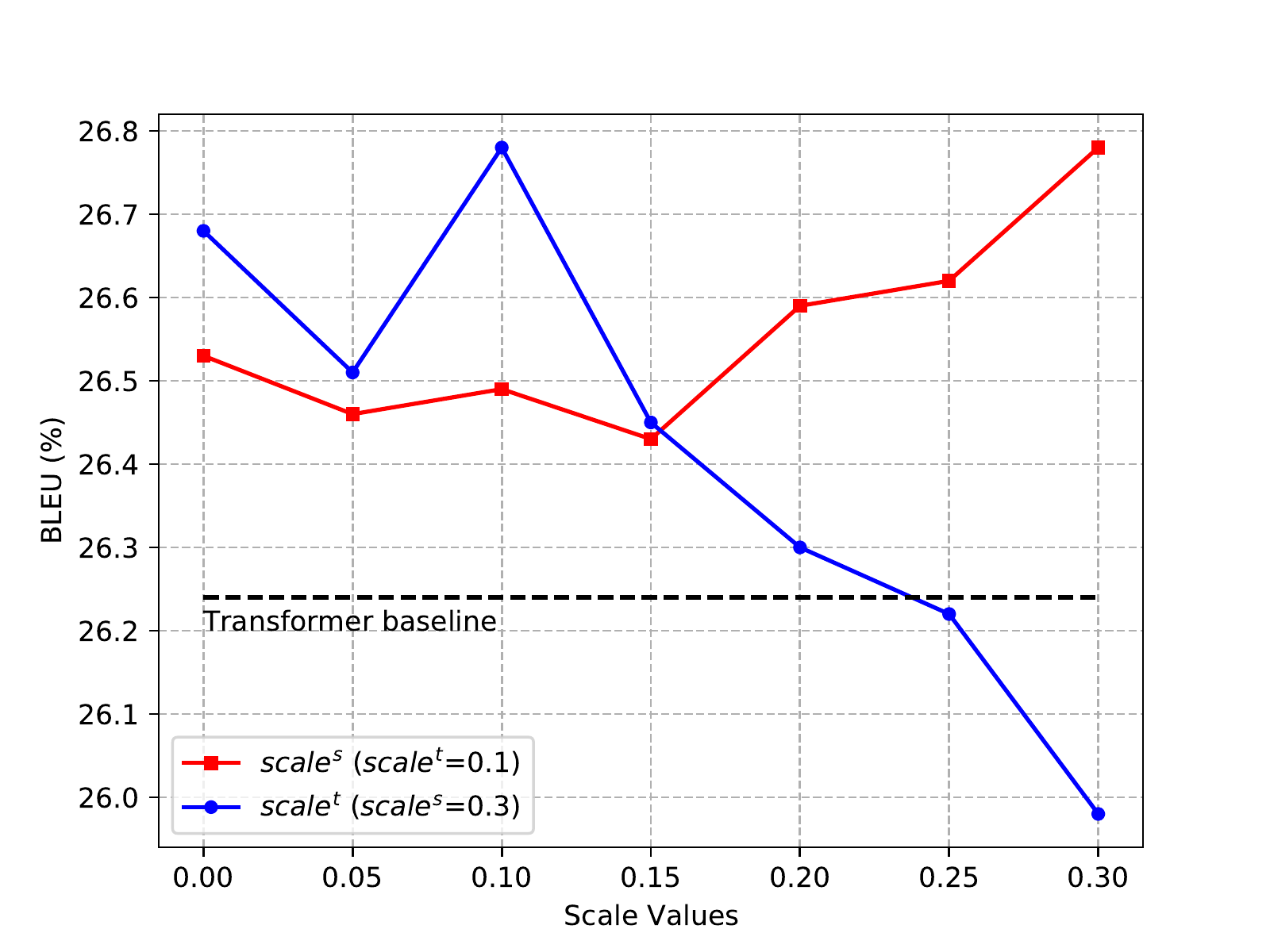}
    \vspace{-5pt}
    \caption{BLEU scores (\%) on the validation set of WMT14 En-De with different $scale^t$ and $scale^s$ that defined in the Equation \eqref{eq:wt} and \eqref{eq:ws}.}
    \label{fig:scale_tune_ende}
    \vspace{-3pt}
\end{figure}

\paragraph{Scale Setting.}
The two hyperparameter $scale^{t}$ and $scale^{s}$ in Equation \eqref{eq:wt} and Equation \eqref{eq:ws} determine the effects of token-level and sentence-level CBMI. To investigate the effects of the two CBMI in different granularities, we firstly fix $scale^{t}$ to a moderate value, {\em i.e.}, 0.1, and tune $scale^{s}$ from 0.0 to 0.3 with the step of 0.05. The detailed results are shown in Figure \ref{fig:scale_tune_ende}. We observe that models perform better with larger $scale^{s}$, which conforms with our conjecture in Section \ref{sec:weight} that well-matched sentence pairs contribute more to NMT models. Then we fix $scale^{s}$ to 0.3 and tune $scale^{t}$ in a similar way. We find it better to keep $scale^{t}$ in a small range and too large value is harmful for models. We conjecture that over-focus on the high-CBMI tokens brings another imbalance for training and may hurt the models. Thus we set $scale^{t}$ to 0.1 in our following experiments. 

% \vspace{-5pt}
\subsection{Baseline Systems} \label{sec:systems}
% \vspace{-5pt}
We implement our approach based on the Transformer \cite{transformer} and compare it with some mainstream adaptive training methods (detailed hyperparameter settings are provided in Appendix \ref{sec:appendix_details}). 
\paragraph{Transformer.} We follow the standard base/big model configurations \cite{transformer} to implement our baseline systems.
\paragraph{Freq-Exponential.} \citet{gu-etal-2020-token} use monolingual token frequency to design an exponential weight function for token-level adaptive training:
\begin{equation*}
    w_j=A\cdot e^{-T\cdot Count(y_j)}+1
\end{equation*}
where $A$ and $T$ are two hyperparameters to adjusting the distribution of weights.
\paragraph{Freq-Chi-Square.} \citet{gu-etal-2020-token} use the chi-square distribution to filter out extremely low frequency target tokens:
\begin{equation*}
    w_j=A\cdot Count(y_j)^2 e^{-T\cdot Count(y_j)}+1
\end{equation*}
where $A$ and $T$ play the same roles as above.
\paragraph{BMI-adaptive.} \citet{xu-etal-2021-bilingual} calculate BMI (in Equation \eqref{eq:bmi}) during the data pre-processing stage and scale it for adaptive loss weights.
\begin{equation}
    w_j=S\cdot {\rm BMI}(\mathbf{x},y_j)+B
\end{equation}
where $S$ and $B$ are hyperparameters to scale BMI to an appropriate range.
\paragraph{Focal Loss.} \citet{focal_loss} propose the focal loss for objective detection tasks to solve the class imbalance problem.
Here we introduce it into NMT.
\begin{equation}
    \mathcal{L}_{fl}=-(1-\alpha p)^\gamma \log p
\end{equation}
where $\alpha$ and $\gamma$ are hyperparameters to adjust the loss weight and $p$ is the NMT predicted probability.
\paragraph{Anti-Focal Loss.} \citet{raunak-etal-2020-long} design an anti-focal loss function to solve the long-tailed problem in NMT by incorporating the inductive bias of inference into training.
\begin{equation}
    \mathcal{L}_{afl}=-(1+\alpha p)^\gamma \log p
\end{equation}
where $\alpha$ and $\gamma$ are similarly as the above focal loss.

% \vspace{-5pt}
\paragraph{Self-Paced Learning.} \citet{wan-etal-2020-self} calculate model confidence  via Monte Carlo dropout sampling \cite{monte_carlo} to measure the token difficulty and use it to re-weight the training losses of tokens.

\paragraph{Simple Fusion.} \citet{stahlberg-etal-2018-simple} propose two simple strategies (i.e., \textsc{PreNorm} and \textsc{PostNorm}) to fuse the NMT probabilities with the LM probablities and directly optimize the fusion during the NMT training process\footnote{The results in Table \ref{tab:main_results} are the higher ones between the two strategies.}.

\paragraph{LM Prior.} \citet{baziotis-etal-2020-language} propose to distill the prior knowledge from LMs trained on rich-resource monolingual data to low-resource NMT models\footnote{We did not use extra monolingual data for the LMs in `Simple Fusion' and `LM Prior' in our implementation for fair comparison.}:
\begin{equation}
    \mathcal{L}_{lmp}=\mathcal{L}_{\rm NMT} + \lambda\cdot\mathcal{L}_{\rm KL}(p_{\rm LM}||p_{\rm NMT};\tau)
\end{equation}
where $\lambda$ weights the distillation term and $\tau$ is the softmax temperature \cite{hinton2015distilling}.
% \vspace{-5pt}
% \paragraph{D2GPo.}\citet{Li2020Data-dependent} design a word-similarity based Gaussian prior for text generation.

% \paragraph{Mixed Cross Entropy.}\citet{pmlr-v139-li21n} use model predictions as additional labels and mix them with the ground-truth labels to form the final labels.

% \paragraph{CBMI-adaptive.} Our CBMI-based multi-granularity weight adjustment strategy described in Section \ref{sec:weight}.
% \vspace{-5pt}
% , {\em i.e.}, models are optimized through Equation \eqref{eq:loss_weight}.

% \paragraph{CBMI-prior selection.}Our CBMI-based prior selection strategy described in Section \ref{sec:selection}.
% % , {\em i.e.}, models are optimized through Equation \eqref{eq:prior_select}.

% \paragraph{CBMI-both weight and prior selection.}The final objective with both CBMI-based weight adjustment and CBMI-based prior selection.

\subsection{Results}
The overall results on two WMT tasks based on the Transformer$_{base}$ and Transformer$_{big}$ configurations are shown in Table \ref{tab:main_results}. 
Under the Transformer$_{base}$ setting, CBMI-based adaptive training can respectively improve +0.91 and +0.85 BLEU scores on En-De and Zh-En tasks compared to the Transformer baseline. 
Compared to the most related yet target-context-free strategy `BMI-adaptive', our CBMI-based adaptive training strategy can respectively yield significant improvements up to +0.46 and +0.44 BLEU scores on En-De and Zh-En, which demonstrate the significance of the target context for token assessment in token-level adaptive training. 
Compared with the best performing baseline `Self-Paced Learning', our approach still outperforms it by +0.32 and +0.46 BLEU scores on the two tasks. 
Our conjecture is that CBMI not only reflects the model competence used in `Self-Paced Learning' but also further incorporates the linguistic statistical information from the target-side LM, thus reflects more explicit translation property ({\em i.e.,} adequacy). 
However, other LM enhanced methods (\textit{e.g.,} `Simple Fusion' and `LM Prior') bring limited improvement or even degradation to the NMT models when there is no extra data for the LMs, which further proves the utilization of the LM in our approach is more effective.  
Under the Transformer$_{big}$ setting, where the performances of existing methods are limited, our method can still bring the improvement of +0.81 and +0.82 BLEU scores on the En-De and Zh-En, which demonstrates the superiority of CBMI under stronger baselines. 

\section{Analysis}

In this section, we provide in-depth analyses on the effectiveness of our CBMI and conduct experiments on the validation set of WMT14 En-De with the Transformer$_{base}$ model.

\subsection{Effects of Different Levels of CBMI}
We take the Transformer$_{base}$ as baseline, and then apply adaptive training based on the token-level CBMI, the sentence-level CBMI, and both of them, respectively.
Results are listed in Table \ref{tab:ablation}. We observe certain improvements (+0.29 and +0.44 BLEU scores) when separately applying the token- and sentence-level CBMI based approaches. It suggests that our CBMI can measure the token importance from different granularities, and up-weight the important tokens or sentence pairs can improve translation quality.
Furthermore, the combination of both the token- and sentence-level CBMI brings further improvement (+0.55 BLEU scores), which illustrates that the CBMI in different granularities are complementary and have cumulative gains.

% Since the final training weight of our approach is determined by both the token-level and sentence-level CBMI, we conduct ablation experiments to investigate the impacts of them and list results in Table \ref{tab:ablation}. 
% By removing the token-level CBMI, the performance decreases by 0.1 BLEU score, indicating that focusing more on the target tokens with higher CBMI in each sentence is indeed beneficial for models. 
% Additionally, we also observe a performance drop of 0.24 BLEU score when we remove the sentence-level CBMI. This drop demonstrates the importance of the sentence-level CBMI in our approach.
% When the two variables are both ablated, the model degenerates to the Transformer baseline and yields a further drop of 0.54 BLEU, which reveals that the information of different granularities are complementary with each other.
\begin{table}[t]
    \centering
    \resizebox{0.95\linewidth}{!}{
    \begin{tabular}{l|l}
        \bottomrule
        \textbf{Model} \qquad\qquad\qquad\qquad\qquad & \textbf{BLEU} \\
        \hline
        Transformer$_{base}$ & 26.24 \\
        \quad+ token-level CBMI & 26.53 (+0.29) \\
        \quad+ sentence-level CBMI &  26.68 (+0.44)\\
        \quad+ token- \& sentence-level CBMI & \textbf{26.78 (+0.54)} \\
        \toprule
    \end{tabular}
    }
    \caption{BLEU scores (\%) of CBMI at different granularities on the validation set of WMT14 En-De.}
    \label{tab:ablation}
    % \vspace{-10pt}
\end{table}

\subsection{Costs of Computing and Storage}
% \vspace{-5pt}
\label{sec:analysis_cost}
In this section, we compare our CBMI-based approach with the BMI-based adaptive training in terms of the number of trainable parameters, the CPU computational costs of pre-processing, the GPU computational costs of training, and disk cost for storing intermediate variables. 
As shown in Table \ref{tab:efficiency}, the vanilla BMI-based approach requires additional 12 CPU hours to obtain the BMI values during the pre-processing stage, and about 2.0 GB of disk space to store these BMI values. 
To make matters worse, the costs of CPU calculation and disk storage will increase dozens of times (approximately equal to the average length of target sentences) when conducting the target-context-aware calculations for BMI.
In contrast, our CBMI-based approach gets rid of the CPU computational costs, and thus has no additional storage overhead. 
Although we introduce an additional LM to calculate the CBMI values, it only brings a slight increase of model parameters and GPU calculation cost during model training. 
Particularly, our proposed method simply modifies the training loss of NMT, and thus has no effect on the inference speed.
In short, our CBMI can be efficiently calculated during model training without any pre-specific statistical calculations and storage overhead, which makes it feasible to supplement target context information for statistical metrics.
\begin{table}[t]
    \centering
    \resizebox{\linewidth}{!}{
    \begin{tabular}{l|cccc}
        \bottomrule
        \multirow{2}*{\textbf{Method}} & \textbf{Pre-process} & \textbf{\#Params} & \textbf{Train} & \textbf{Disk} \\
        ~ & (hour) & (M) & (hour) & (GB) \\
        \hline
        Transformer$_{base}$ & 0  & 65  & 10 & 0\\
        \quad + BMI & 12 & 65  & 11 & 2.0 \\
        \quad\quad + target context & $\approx$ 12$\times$N & 65  & -- & $\approx$ 2.0$\times$N  \\
        \quad + CBMI & 0  & 101 & 12 & 0 \\
        \toprule
    \end{tabular}
    }
    % \vspace{-4pt}
    \caption{
    The costs of calculation and storage of the BMI- and CBMI-based approaches on the WMT14 En-De (100k training steps). `N' refers to the average length of target sentences.}
    \label{tab:efficiency}
    % \vspace{-3pt}
\end{table}
\subsection{Human Evaluation}
\begin{table}[t]
    \centering
    \resizebox{0.95\linewidth}{!}{
    \begin{tabular}{l|ccc}
        \bottomrule
        \textbf{Model} & \textbf{Adequacy} & \textbf{Fluency} & \textbf{Avg.} \\
        \hline
        Transformer$_{base}$ & 4.25 & 4.69 & 4.47 \\
        % \hline
        \quad + CBMI-adaptive & \textbf{\ \,4.53$^\ast$} & \textbf{4.75} & \textbf{4.64}\\
        \toprule
    \end{tabular}}
    % \vspace{-2pt}
    \caption{Human evaluation on adequacy and fluency. $\ast$ means the average Cohen's Kappa \cite{cohen1960coefficient} is higher than 0.6, which indicates substantial agreement between three annotators \cite{landis1977measurement}.}
    \label{tab:human_evaluation}
    % \vspace{-3pt}
\end{table}

To verify whether our CBMI measurement is indeed highly related to the translation adequacy of NMT models, as we conjectured in Section \ref{sec:weight}, we conduct the human evaluation in terms of adequacy and fluency.
We randomly sample 100 sentences from the test set of WMT19 Zh-En and invite three annotators to evaluate the translation adequacy and fluency. Scores for both indexes are limited in [1,5]. For adequacy, `1' represents irrelevant to the source sentence and `5' represents semantically equal. For fluency, `1' means unintelligible and `5' means fluent and native. We finally average the scores from three annotators and list the results in Table \ref{tab:human_evaluation}.
% As we conjectured earlier, 
We observe that our approach significantly promotes the translation adequacy of the Transformer$_{base}$ baseline, and meanwhile slightly promotes the translation fluency. 
It indicates that the CBMI measurement is highly related to the adequacy of NMT models, and focusing more on the tokens with high CBMI can improve translation adequacy, and thus improve translation performance.
% \vspace{-5pt}
\subsection{Prior Selection based on CBMI}
Given that CBMI reflects the dependency between a target token and its source sentence on the condition of its target context, in this section, we explore whether CBMI can serve as an indicator for selecting an appropriate prior distribution to improve the NMT model.  
Prior distributions have been proved for their ability to provide additional knowledge for models \cite{baziotis-etal-2020-language,Li2020Data-dependent}. 
Thus we try three generated distributions as prior distributions for NMT models, {\em i.e.}, the translation model distribution (TM prior), the language model distribution (LM prior), and the softmax normalized CBMI distribution (CBMI prior). 
% The first two distributions are both output by models and the last is the softmax normalization of CBMI values calculated by the former ones. 

\begin{figure}[t]
    \centering
    % \resizebox{0.88\linewidth}{!}{
    \includegraphics[width=0.85\linewidth]{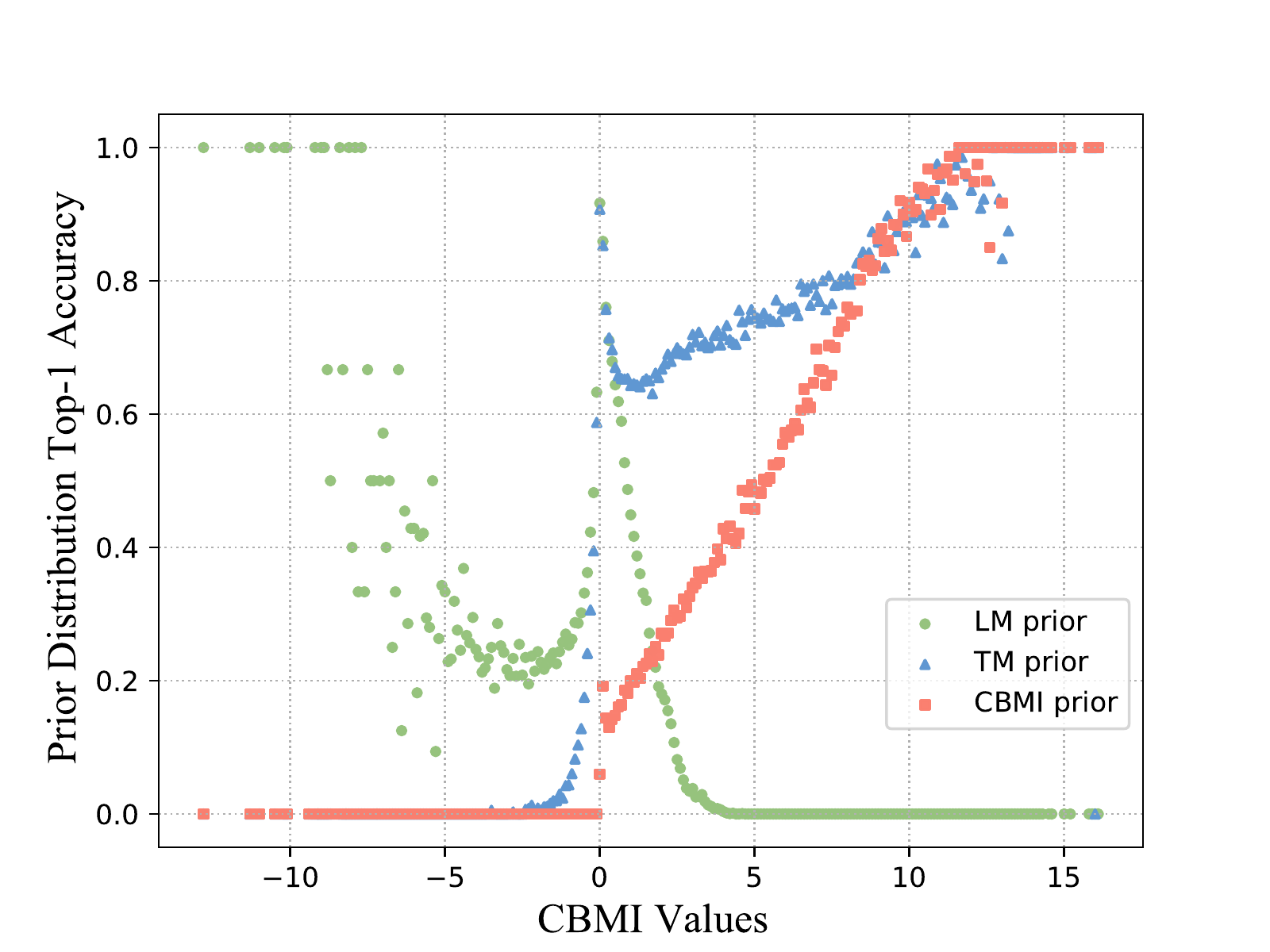}
    % \vspace{-2pt}
    \caption{The most accurate prior distribution for tokens with different CBMI values. The LM prior (green circles) performs most accurate for tokens with lower CBMI values, the TM prior (blue triangles) performs best for tokens with moderate CBMI values, and the CBMI prior (red squares) performs best for tokens with higher CBMI values.}
    % }
    \label{fig:prior_acc}
    % \vspace{-3pt}
\end{figure}

To verify the correctness of these prior distributions, we firstly calculate the top-1 accuracies of these distributions according to different tokens and surprisingly observe that the accuracies are highly related to the CBMI values of tokens. 
As shown in Figure \ref{fig:prior_acc}, the most accurate prior for target tokens with different CBMI values is not always consistent.
Based on this observation, we further design a CBMI-based prior selection strategy to choose the best prior distribution for each token. 
The details of the selection strategy are seen in Appendix \ref{sec:selection_detail}.

As shown in Table \ref{tab:prior}, all these prior distributions can provide helpful guidance and enhance the baseline model. 
More importantly, the CBMI-based prior selection strategy can achieve a better performance compared with the single prior, demonstrating that CBMI also serves as an appropriate indicator for the translation prior selection. We will explore the more sophisticated CBMI-based prior selection strategy in the future work.

\begin{table}[t]
    \centering
    \resizebox{0.8\linewidth}{!}{
    \begin{tabular}{l|l}
        \bottomrule
        \textbf{Prior Distribution} \qquad\qquad & \textbf{BLEU} \\
        \hline
        Transformer$_{base}$ & 26.24\\
        \quad + LM Prior & 26.73 (+0.49)\\
        \quad + TM Prior & 26.61 (+0.37)\\
        \quad + CBMI Prior & 26.57 (+0.33)\\
        \quad + Prior Selection & \textbf{26.75 (+0.51)} \\
        \toprule
    \end{tabular}
    }
    % \vspace{-5pt}
    \caption{BLEU scores (\%) on WMT14 En-De validation set for different prior distributions on all tokens.}
    \label{tab:prior}
    % \vspace{-5pt}
\end{table}

% \iffalse
\section{Related Work}
\paragraph{Language Model Enhanced NMT.}Exploiting the information in language models is a common solution to improve NMT models. 
In low-resource scenarios, LMs trained on extra monolingual data are usually more informative and thus used to fuse with NMT prediction \cite{gulcehre2015using,integrate-lm-2017,cold_fusion,stahlberg-etal-2018-simple}, provide prior knowledge for NMT models \cite{baziotis-etal-2020-language} and enhance representations of NMT \cite{use-bert-for-nmt,bert-nmt}.
In data augmentation methods, LMs are also widely used to generate contextual substitutions of words in sentences \cite{kobayashi-2018-contextual,wu2018conditional,gao-etal-2019-soft}.
Differently, all the aforementioned methods rely on the LMs that are trained on extra data, while the LM in our method does not require extra data and also has no influence on the inference speed.

\section{Conclusion}
% \vspace{-2pt}
In this paper, we propose a target-context-aware metric for target tokens, named conditional bilingual mutual information (CBMI).
% which measures the dependencies between target tokens and their source sentences on the condition of target contexts. 
Compared with previous statistical metrics, our CBMI only increases limited computational costs to incorporate the target context and provides a more suitable assessment for tokens. 
Furthermore, based on the token- and sentence-level CBMI, we design a CBMI-based adaptive training strategy to amply the contributions of the important tokens. 
Experimental results on two WMT tasks demonstrate the effectiveness of our proposed approach.
Further analyses show that CBMI can improve translation adequacy and serve as an appropriate indicator for the translation prior selection.

\section*{Acknowledgements}
The research work descried in this paper has been supported by the National Key R\&D Program of China (2020AAA0108001) and the National Nature Science Foundation of China (No. 61976016, 61976015, and 61876198). The authors would like to thank the anonymous reviewers for their valuable comments and suggestions to improve this paper.

% Entries for the entire Anthology, followed by custom entries
\bibliography{anthology,custom}
\bibliographystyle{acl_natbib}

\appendix

\begin{table*}[t] 
\centering
\begin{tabular}{lcc}
\toprule
\textbf{Model} \qquad\qquad\qquad\qquad\qquad\qquad\qquad & \textbf{WMT14 En$\rightarrow$De} & \textbf{WMT19 Zh$\rightarrow$En}\\
\hline
\hline
Transformer$_{base}$ \cite{transformer} & 28.10 ± 0.20 &  25.36 ± 0.19\\
\quad + Freq-Exponential \cite{gu-etal-2020-token} & 28.43 ± 0.03  & 24.99 ± 0.01 \\
\quad + Freq-Chi-Square \cite{gu-etal-2020-token} & 28.47 ± 0.24 & 25.43 ± 0.72\\
\quad + BMI-adaptive \cite{xu-etal-2021-bilingual} & 28.56 ± 0.09 & 25.77 ± 0.04 \\
\quad + Focal Loss \cite{focal_loss} & 28.43 ± 0.10 & 25.37 ± 0.25\\
\quad + Anti-Focal Loss \cite{raunak-etal-2020-long} & 28.65 ± 0.13 & 25.50 ± 0.33\\
\quad + Self-Paced Learning \cite{wan-etal-2020-self} & 28.69 ± 0.22 & 25.75 ± 0.25 \\
\quad + Simple Fusion \cite{stahlberg-etal-2018-simple} & 27.82 ± 0.17 & 23.91 ± 0.22 \\
\quad + LM Prior \cite{baziotis-etal-2020-language} & 28.27 ± 0.10 & 25.71 ± 0.42 \\
% \quad + D2GPo \cite{Li2020Data-dependent} & 28.24 ± 0.08 \\
% \quad + Mixed Cross Entropy \cite{pmlr-v139-li21n} & 28.24 ± 0.29 \\
\quad + CBMI-weight (ours) & \textbf{\ \ \ 29.01 ± 0.08$^{\ast\ast}$} & \textbf{\ \ \ 26.21 ± 0.30$^{\ast\ast}$} \\
% \quad + CBMI-prior selection (ours) & 28.48 ± 0.04$\ast$ \\
% \quad + CBMI-both weight and prior selection (ours) & \textbf{28.94 ± 0.07$\ast$} \\
\midrule
Transformer$_{big}$ \cite{transformer} & 29.31 ± 0.29 & 25.48 ± 0.31  \\
\quad + Freq-Exponential \cite{gu-etal-2020-token} & 29.66 ± 0.04 & 25.57 ± 0.15\\
\quad + Freq-Chi-Square \cite{gu-etal-2020-token} & 29.64 ± 0.45 & 25.64 ± 0.23\\
\quad + BMI-adaptive \cite{xu-etal-2021-bilingual} & 29.69 ± 0.15 & 25.81 ± 0.13 \\
\quad + Focal Loss \cite{focal_loss} & 29.65 ± 0.11 & 25.54 ± 0.09 \\
\quad + Anti-Focal Loss \cite{raunak-etal-2020-long} & 29.72 ± 0.16 & 25.64 ± 0.18 \\
\quad + Self-Paced Learning \cite{wan-etal-2020-self} & 29.85 ± 0.18 & 25.88 ± 0.23\\
% \quad + Simple Fusion \cite{stahlberg-etal-2018-simple} & & \\
% \quad + D2GPo \cite{Li2020Data-dependent} &  \\
% \quad + Mixed Cross Entropy \cite{pmlr-v139-li21n} &  \\
\quad + CBMI-weight (ours) & \textbf{\ \,30.12 ± 0.13$^{\ast}$} & \textbf{\ \,26.30 ± 0.26$^{\ast}$} \\
% \quad + CBMI-prior selection (ours)  \\
% \quad + CBMI-both weight and prior selection (ours)  \\
\bottomrule
\end{tabular}
\caption{The complete results of Table \ref{tab:main_results} containing mean values and standard deviations of BLEU scores.}
\label{tab:complete}
\end{table*}

\section{Complete Results}
\label{sec:full}

To prove the generality of the experimental results, we provide the complete results on two translation tasks which contain mean values and standard deviations in Table \ref{tab:complete}.

\section{Training Hyperparameters and Model Configurations} \label{sec:appendix_config}
To assure the reproducibility of our experimental results, we provide the training details of our method and the model configurations in Table \ref{tab:detail}. 
The NMT models and LMs in our method use the same corpus and BPE vocabulary, so that they can generate two corresponding probability distributions for the same token and calculate its CBMI during training. 
Our LMs have the same model architecture and configuration with the NMT models' decoder except for the cross-attention module, yet we do not share their embedding layers for higher performance.
\begin{table}[H]
    \centering
    \resizebox{\linewidth}{!}{
    \begin{tabular}{l|c|c}
        \bottomrule
        \textbf{Hyperparameters} & \textbf{Base} & \textbf{Big} \\
        \hline
        Embedding Size & 512 & 1024 \\
        Encoder Layers & 6 & 6 \\
        Decoder Layers & 6 & 6 \\
        Attention Heads & 8 & 16 \\
        LM Layers & 6 & 6 \\
        LM Attention Heads & 8 & 16 \\
        Residual Dropout & 0.1 & 0.3 \\
        Attention Dropout & 0.1 & 0.1 \\
        Activation Dropout & 0.1 & 0.1 \\
        Learning Rate & 7e-4 & 5e-4 \\
        Learning Rate Decay & inverse sqrt & inverse sqrt \\
        Warmup Steps & 4000 & 4000 \\
        Layer Normalization & PostNorm & PostNorm \\
        \toprule
    \end{tabular}
    }
    \caption{Training hyperparameters and model configurations of our method.}
    \label{tab:detail}
\end{table}

\section{Implementation Details for Baseline Systems}
\label{sec:appendix_details}
To make our experimental comparison more convincing, we present the details of hyperparameters involved in the baseline systems described in Section \ref{sec:systems}.
\paragraph{Freq-Exponential.} Following the best hyperparameter setting in \cite{gu-etal-2020-token}, we set A to 1.0 and T to 1.75 for the En-De task, and A to 1.0 and T to 0.35 for the Zh-En task.

\paragraph{Freq-Chi-Square.} Similarly, we set A to 1.0 and T to 2.50 for the En-De task, and A to 1.0 and T to 1.75 for the Zh-En task according to \cite{gu-etal-2020-token}.

\paragraph{BMI-adaptive.} According to the settings in  \cite{xu-etal-2021-bilingual}, we set S to 0.15 and B to 0.8 for the En-De task and S to 0.1 and B to 1.0 for the Zh-En task.

\paragraph{Focal Loss.} As suggested in \cite{focal_loss}, we fix $\gamma$ to 1.0 and search a $\alpha$ among [0.1, 0.5] which performs best on the validation sets of two tasks. Finally, we set $\alpha$ to 0.1 for both tasks.

\paragraph{Anti-Focal Loss.} Similar with the settings in focal loss, we also fix $\gamma$ to 1.0 and tune $\alpha$ for two tasks. Lastly, we also set $\alpha$ to 0.1 for both tasks.

\paragraph{LM Prior.} We set the softmax temperature $\tau$ to 2.0 following the settings in \cite{baziotis-etal-2020-language} while $\lambda$ to 0.1 according to the performances on the validation sets.

\section{Details for the Prior Selection Strategy}
\label{sec:selection_detail}

\begin{figure}[t]
    \centering
    \includegraphics[width=\linewidth]{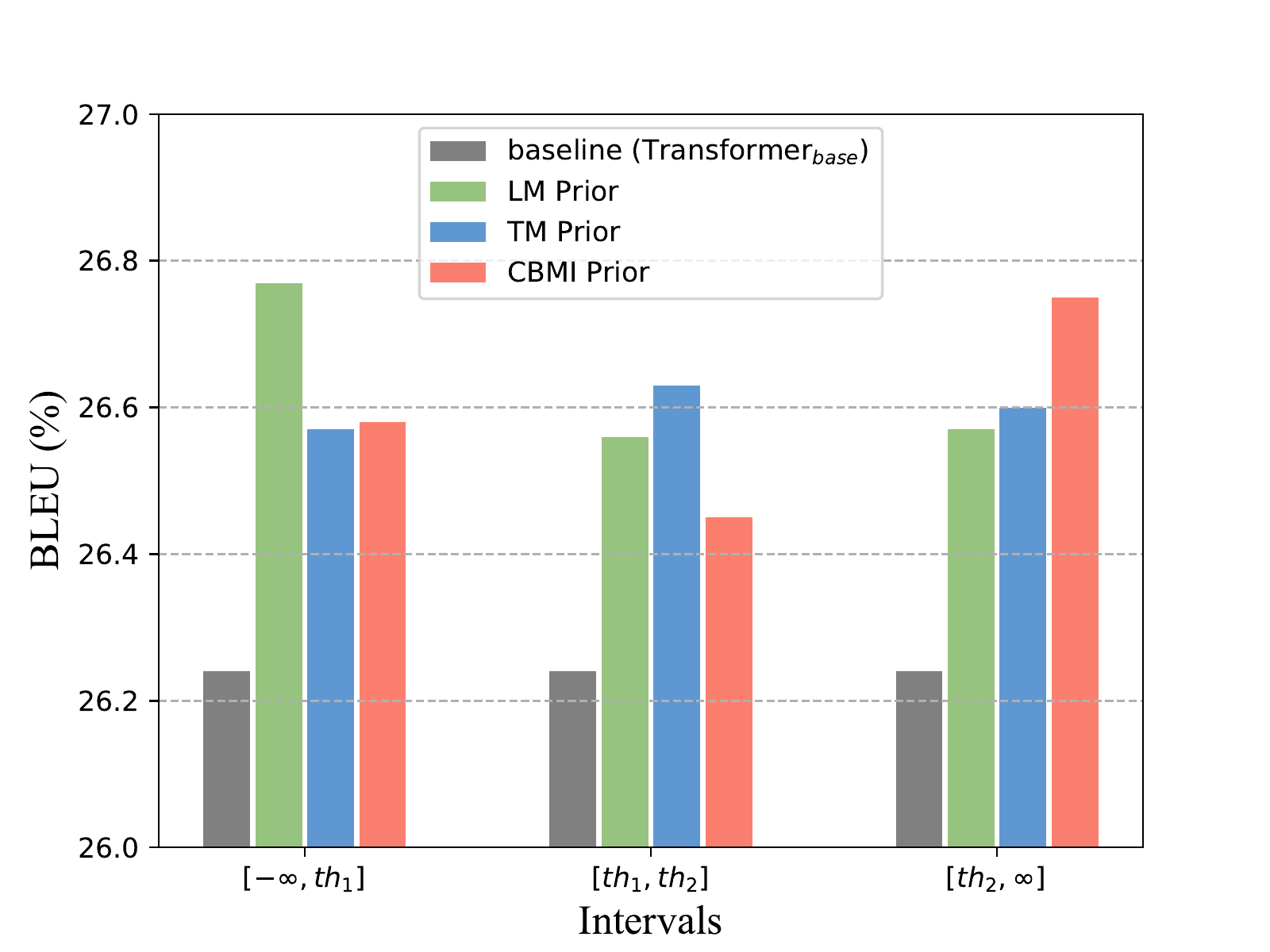}
    \caption{BLEU scores on the validation set of WMT14 En-De for different prior distributions on different CBMI intervals.}
    \label{fig:selection}
\end{figure}

In our prior selection strategy, we firstly divide the target tokens in each mini-batch into three intervals according to their original CBMI values. Corresponding to the observation in Figure \ref{fig:prior_acc}, we respectively apply the LM prior, the TM prior and the CBMI prior on the tokens in the three intervals. Formally, the prior distribution $q(y_j)$ for target token $y_j$ can be represented as follows:
\begin{equation} \label{eq:select}
    \resizebox{0.86\linewidth}{!}{$
        q(y_j)=\left\{
        \begin{array}{ll}
            q_{\rm LM}, & {\rm CBMI}(\mathbf{x};y_j) \in [-\infty, th_1] \\
            q_{\rm TM}, & {\rm CBMI}(\mathbf{x};y_j) \in [th_1, th_2] \\
            q_{\rm CBMI}, & {\rm CBMI}(\mathbf{x};y_j) \in [th_2, \infty]
        \end{array}
        \right.$}
\end{equation}
where $th_1$ and $th_2$ are two hyperparameters and empirically set to 0 and 8 according to the observations in Figure \ref{fig:prior_acc}. $q_{\rm LM}$, $q_{\rm TM}$, $q_{\rm CBMI}$ represent the aforementioned three prior distributions.

Subsequently, we calculate the cross-entropy loss between the selected prior distribution and the model predicted distribution as an additional term and incorporate it with the original cross-entropy loss in Equation \eqref{eq:ce_loss} to make up the new training objective:
\begin{align}
    \label{eq: selection}
    \mathcal{L}(\theta)=&\mathcal{L}_{\rm CE}(\theta) \nonumber \\
    &+ \lambda \cdot \sum_{y=1}^{|\mathbf{y}|}-q(y_j)\log p(y_j|\mathbf{y}_{<j},\mathbf{x};\theta)
\end{align}
where $\lambda$ is a hyperparameter that controls the effect of prior distribution. In our experiments, we set $\lambda$ to 0.1 according to the performances on the validation set.

To verify the reasonablility of the prior selection strategy, we compare the effects of the three priors on each single CBMI intervals in Figure \ref{fig:selection}. 
As we expected, the BLEU results also conform with the accuracy results in Figure \ref{fig:prior_acc}, indicating that the most helpful prior distribution can be highly related to the CBMI values of tokens.

\end{document}